**LLMs for energy and macronutrients estimation using only text data from 24-hour dietary recalls: a parameter-efficient fine-tuning experiment using a 10-shot prompt**

Rodrigo M Carrillo-Larco[1,2]

1. Hubert Department of Global Health, Rollins School of Public Health, Emory University, Atlanta, GA, USA.
2. Emory Global Diabetes Research Center of Woodruff Health Sciences Center, Emory University, Atlanta, USA

**CORRESPONDING AUTHOR**

Rodrigo M Carrillo-Larco, MD, PhD

Rollins School of Public Health, Emory University, Atlanta, GA, USA.

rmcarri@emory.edu

**ABSTRACT**

**BACKGROUND**: Most artificial intelligence tools used to estimate nutritional content rely on image input. However, whether large language models (LLMs) can accurately predict nutritional values based solely on text descriptions of foods consumed remains unknown. If effective, this approach could enable simpler dietary monitoring without the need for photographs. **METHODS**: We used 24-hour dietary recalls from adolescents aged 12–19 years in the National Health and Nutrition Examination Survey (NHANES). An open-source quantized LLM was prompted using a 10-shot, chain-of-thought approach to estimate energy and five macronutrients based solely on text strings listing foods and their quantities. We then applied parameter-efficient fine-tuning (PEFT) to evaluate whether predictive accuracy improved. NHANES-calculated values served as the ground truth for energy, proteins, carbohydrates, total sugar, dietary fiber and total fat. **RESULTS**: In a pooled dataset of 11,281 adolescents (49.9% male, mean age 15.4 years), the vanilla LLM yielded poor predictions. The mean absolute error (MAE) was 652.08 for energy and the Lin's CCC <0.46 across endpoints. In contrast, the fine-tuned model performed substantially better, with energy MAEs ranging from 171.34 to 190.99 across subsets, and Lin's CCC exceeding 0.89 for all outcomes. **CONCLUSIONS**: When prompted using a chain-of-thought approach and fine-tuned with PEFT, open-source LLMs exposed solely to text input can accurately predict energy and macronutrient values from 24-hour dietary recalls. This approach holds promise for low-burden, text-based dietary monitoring tools.

**INTRODUCTION**

Technologies, including artificial intelligence (AI), are increasingly being leveraged for health-related applications, including the areas of nutrition monitoring and dietary assessment.[1-5] Among these, AI tools that use image analysis have been widely adopted to help individuals monitor their food intake.[6-11] For instance, several mobile applications allow users to take photos of their meals and receive feedback on estimated energy and nutritional content.[6-11] Although the accuracy of these applications varies, they are generally considered reliable enough for pragmatic use.[6-11]

With the rapid rise of generative AI—most notably large language models (LLMs), which now engage millions of users daily—the analysis of text-based data has expanded substantially. In the field of nutrition, LLMs capable of processing both text and images have begun to be explored as tools for estimating energy and nutrient intake.[12-15] However, it remains unknown whether LLMs that receive *only* text input—without images—can perform this task accurately. Texting what one eats throughout the day, in the form of a dietary diary, may represent a simpler and less resource-intensive alternative to photographing meals. Moreover, given the widespread use of messaging apps and social media platforms, text-based dietary logging may align more naturally with users' daily behaviors.

In this study, we aimed to evaluate whether an LLM, provided solely with a string of text listing all foods and their respective quantities as collected in a 24-hour dietary recall, could accurately estimate energy and macronutrient intake. We also investigated whether fine-tuning the LLM would enhance its predictive performance.

**METHODS**

**Data sources**

We used data from the National Health and Nutrition Examination Survey (NHANES) in the United States (US).[16,17] NHANES samples non-institutionalized individuals across the country and is designed to produce health estimates representative at the national level. We pooled all consecutive NHANES cycles from 2003-2004 to 2021-2023. Earlier cycles were excluded because they included only a single 24-hour dietary recall, whereas the selected cycles included two non-consecutive 24-hour dietary recalls. All NHANES cycles follow a consistent protocol, including standardized methods for dietary data collection.[16,17]

**Sample**

We included only participants with high-quality 24-hour dietary recalls, as determined by NHANES protocols.[16,17] Individuals who were breastfeeding at the time of data collection were excluded. This analysis included only adolescents aged 12 to 19 years, inclusive. Younger participants were excluded because their dietary recall was completed by a proxy or in the presence of a proxy.[16,17] We also excluded adults, as adolescents represent a distinct population for this experiment. Adolescents frequently interact with mobile devices via text, and most prior research using novel technologies for dietary assessment (e.g., imaging-based nutrient analysis) has focused on adults.[6-11] Thus, adolescents provide a novel and relevant population to assess whether LLMs can accurately estimate energy and macronutrient intake.

**Input data for the LLM**

The pooled dataset of adolescents aged 12-19 years with two high-quality 24-hour dietary recalls and who were not breastfeeding included 24,111 observations. From this pooled dataset, we randomly selected 10 records to construct the 10-shot prompt (Supplementary Table 1). We then used only data from the second day of dietary recall, which included 11,281 unique participants. We chose the second day to reduce computational demands, as using both days (24,111 observations) or only the first day (12,830 observations) would require more computational resources without proportionate added value for our primary objective—to assess the accuracy of LLM estimates for energy and macronutrient intake. Finally, to facilitate processing, we divided the dataset from the second day (11,281 participants) into 10 subsets: one containing 1,129 participants and nine others with 1,128 each. These data subsets were used at different stages of the analytical approach.

**Exposure variable**

The exposure information—later formatted as input data for the LLM—consisted of dietary data provided by NHANES participants during the 24-hour dietary recalls.[16,17] The first recall was conducted in person at a mobile examination center, while the second was conducted by telephone three to five days later, on a different day of the week.[16,17] NHANES nutritional datasets include USDA food codes, each of which can be mapped to a corresponding text descriptor (e.g., food code 11000000 maps to "MILK, HUMAN"). For each participant, all reported food items were combined into a single string, with each item paired with the quantity consumed. This text string, listing all foods and their respective amounts, was presented to the LLM and had the following form: "*PORK CHOP, BREADED, FRIED, LEAN ONLY (22); CHICKEN PATTY/FILLET/TENDERS, BREADED, COOKED (48); BREAD, WHITE (26); SALTY SNACKS, CORN OR CORNMEAL, CORN PUFFS, TWISTS (14); ORANGE, RAW (96); MUSTARD SAUCE (10); TAFFY (15.6); SOFT DRINK, FRUIT-FLAVORED, CAFFEINE FREE (314.68); FRUIT-FLAVORED DRINK, FROM SWEETENED POWDER, FORTIFIED WITH VITAMIN C (203.13).*" Values in parentheses represent the amount consumed (grams). Aside from formatting the food entries as described, we did not alter the NHANES data in any way.

**Outcome variable**

The outcomes of interest—the variables the LLM was tasked with predicting—included energy and five macronutrients. Specifically, these were: energy (kilocalories; NHANES variable DRx1KCAL), protein (grams; DRxIPROT), carbohydrates (grams; DRxICARB), total sugars (grams; DRxISUGR), dietary fiber (grams; DRxIFIBE), and total fat (grams; DRxITFAT). These outcomes were selected because they are commonly targeted in dietary tracking and weight management applications. We used the NHANES data exactly as provided, without any additional processing or modification to the energy and macronutrient variables. These outcome variables were treated as the ground truth and served as the reference standard against which the LLM's predictions were benchmarked.

**Analytical approach**

We conducted two experiments. First, in the baseline LLM evaluation, we used the second data subset (n = 1,128) to prompt the LLM to estimate energy and macronutrient values based solely on the food list, using the 10-shot prompt. Notably, the 10-shot prompt was structured using a chain-of-thought approach, in which each step of the reasoning process was explicitly articulated to guide the LLM through the task. Second, in the fine-tuned LLM evaluation, we applied Parameter-Efficient Fine-Tuning (PEFT) to the LLM using the first data subset (n = 1,129) and then used the fine-tuned model to generate predictions across the remaining data subsets (nine data subsets; n = 1,128 each). In summary, the first subset (n = 1,129) was used exclusively for PEFT, the second subset (n = 1,128) was used to evaluate both the vanilla and fine-tuned LLM, and the remaining subsets (n = 1,128 each) were used to evaluate the fine-tuned LLM.

PEFT was conducted using the Unsloth ecosystem,[18] which enables fast and efficient fine-tuning of quantized LLMs. We ran PEFT for 10 epochs on the first data subset using the 10-shot prompt. Among the models available in Unsloth,[18] we used the Mistral-Small-24B-Instruct-2501 (4-bit), as preliminary testing showed that other LLMs failed to return output in the desired format.

To assess the accuracy of the LLM predictions—both from the vanilla model (before fine-tuning) and the fine-tuned model—we compared them against the ground truth values (i.e., energy and macronutrients reported in the NHANES datasets). We computed the following metrics: mean squared error (MSE), mean absolute error (MAE), mean absolute percentage error (MAPE), root mean squared error (RMSE), $R^2$, and Lin's concordance correlation coefficient (Lin's CCC). We also computed paired t-tests. Finally, we generated Bland–Altman plots. Ideally, the MSE, MAE, MAPE, and RMSE should be close to zero, indicating minimal error between the ground truth and the predictions. In contrast, $R^2$ and Lin's CCC should be close to 1, reflecting strong concordance between the predicted and observed values. The Bland-Altman plots provide a visual representation of the differences between the ground truth and the predictions, where most values should fall within the limits of agreement (green dotted lines), and the mean difference (red solid line) should be close to zero indicating minimal bias between the ground truth and the predictions.

In a few instances, the LLM returned predictions that were not in the expected format and, therefore, these cases were excluded from the computation of validation metrics. In such cases, the model failed to respond as instructed. The effective sample size used to compute the validation metrics is reported and was always >89% of the respective data subset.

The analysis was conducted using Python, and the analysis code is provided as Supplementary Materials. We did not account for the complex survey design of the NHANES (e.g., primary sampling units or sampling weights), because our goal was to compare the ground truth values with the LLM-based predictions rather than to produce nationally representative population-weighted estimates.

**Ethical considerations**

We analyzed publicly available data that contain no personal identifiers or sensitive information. Human subjects were not directly involved in this research. Ethical approval was not sought as the study posed minimal risk. NHANES participants provided informed consent.[16,17]

**Role of the funding source**

No specific funding supported this work.

**Public involvement**

Neither members of the public nor patients were involved in the design, analysis, or interpretation of this study.

**RESULTS**

**Study sample**

In the pooled dataset including only the second day of 24-hour dietary recalls (n = 11,281), the sample was evenly split by sex, with 49.9% male and 50.1% female participants. The mean age was 15.4 years, and the median age was 15.0 years. A descriptive summary of the observed outcomes (energy and macronutrients) is presented in Table 1.

**Predictions – before fine-tuning**

The vanilla model—i.e., the LLM using the 10-shot prompt without fine-tuning—produced poor predictive performance. The MAE was highest for energy (652.08) and lowest for dietary fiber (4.90; Table 2). Lin's CCC was consistently low, with values of 0.463 for energy and as low as 0.113 for total fat. Bland-Altman plots also indicated poor agreement between predicted and observed values, with a tendency for error magnitude to increase at higher levels of energy and macronutrient intake (Supplementary Figure 1).

**Predictions – after fine-tuning**

Predictions improved substantially after PEFT (Table 2). This improvement was consistent across all nine test data subsets. Energy remained the outcome with the highest MAE, ranging from 171.34 (data subset 5) to 190.99 (data subset 9). Fiber consistently showed the lowest error, with MAEs ranging from 1.94 (data subset 2) to 2.60 (data subset 8). Lin's CCC exceeded 0.91 for all outcomes across all data subsets, with a few exceptions. For total fat, Lin's CCC ranged from 0.831 (data subset 5) to 0.916 (data subset 2). For fiber, the coefficient dropped to 0.874-0.878 in two subsets (data subsets 9 and 10). For total sugars, the lowest Lin's CCC was 0.896 in data subset 9. Bland-Altman plots showed good agreement between predicted and reference values, with no evident systematic bias across the range of measurements (Supplementary Figures 2-10).

## DISCUSSION

**Main findings**

Using a 10-shot chain-of-thought prompt and an open-source LLM presented with text inputs—no images—describing foods and their respective quantities, we found that the LLM in its baseline form did not produce accurate predictions for energy and five macronutrients (protein, carbohydrates, sugars, fiber, and fat). However, after PEFT, the LLM's predictive performance improved substantially. For instance, the fine-tuned LLM showed strong agreement with reference values, with Lin's CCC exceeding 89% for all outcomes across nine independent data subsets.

These findings suggest that open-source LLMs, when fine-tuned using high-quality 24-hour dietary recall data, can accurately estimate energy and macronutrient content from text alone—without requiring accompanying food images. This capability has potential applications for facilitating dietary data collection in research settings, as well as for enhancing real-time monitoring and personalized dietary feedback among individuals aiming to manage their weight or improve nutritional habits.

**Strengths and limitations**

This study has important strengths. We leveraged a large, nationally representative dataset from the NHANES,[16,17] spanning multiple years. NHANES is widely regarded as a gold standard in population-based health research, with high-quality data collected using rigorous protocols.[16,17] We relied on 24-hour dietary recalls administered by trained professionals, and the outcome variables—energy and macronutrients—were derived using state-of-the-art, validated methods.[16,17] In addition, we used an open-source LLM and PEFT,[18] enhancing the accessibility, reproducibility, and scalability of our approach using emerging technologies.

However, limitations must be acknowledged. First, the input data from NHANES had already undergone preprocessing, which may not reflect the complexity or variability of real-world free text inputs. Future work should evaluate LLM performance using free-text dietary descriptions provided by lay individuals in everyday contexts—similar to how one might text a friend or report to a health worker what they ate the previous day. Second, we limited our analysis to English-language inputs. While many LLMs are multilingual, their performance may vary depending on the volume and quality of non-English data included in their pretraining.[19] Thus, the accuracy reported in this study may not generalize to inputs in other languages. Future studies should explore multilingual performance and consider cultural or regional food naming differences.[14] Third, we focused exclusively on energy and macronutrients—nutrients most commonly tracked in diet-monitoring apps and interventions for weight management. However, other dietary components, such as micronutrients (e.g., sodium, calcium, vitamin D), are important for specific subpopulations. For instance, salt intake is particularly relevant for individuals with hypertension, while iron and folate are crucial for pregnant women. Future research should assess LLMs' ability to estimate a broader array of dietary constituents. Fourth, we relied on a single fine-tuned model applied to 24-hour dietary recall data, which may limit generalizability. Other dietary assessment tools (e.g., food frequency questionnaires) might pose unique challenges to LLM performance and should be tested separately. Lastly, although PEFT is computationally efficient,[18] fully fine-tuning the model—or training a new LLM exclusively on dietary data—may yield much better predictive performance.

**Implications**

This study demonstrates that LLMs can accurately predict energy and macronutrient content when provided with a plain-text string listing food items and their consumed quantities. These findings expand upon existing literature,[1-5] which has largely focused on image-based dietary assessment.[6-15] While recent applications of LLMs in nutrition have harnessed their ability to process visual inputs,[12-15] few or none have investigated the potential of text-only inputs for nutritional predictions. For instance, Haman *et al.*[20] used an OpenAI LLM to estimate the nutritional content of 236 individual food items, rather than complete 24-hour dietary recalls.

Pending external validation and further refinement to address the limitations discussed, our findings offer preliminary yet compelling evidence that LLMs can serve as effective tools for dietary monitoring. Future applications, ideally tested in trials, should evaluate whether mobile apps that accept only text input can deliver accurate nutritional assessments. These assessments could then be leveraged to trigger tailored or personalized dietary recommendations, complementing the growing literature on how LLMs can aid in personalized dietary prescription.[20-25] This also aligns with the increasing interest in precision nutrition.[26,27] In parallel with the widespread use of images in nutrition tracking,[6-11] text-based LLM solutions may represent a simpler and more accessible alternative, especially in settings where image capture is impractical.

Beyond personal health applications, our findings may also hold implications for research and public health nutrition. The collection and analysis of 24-hour dietary recalls are typically labor- and time-intensive processes requiring trained interviewers and specialized coding. Our results suggest that, with appropriate fine-tuning, LLMs could automate and streamline this process—receiving plain-text dietary inputs and instantly returning estimates of energy and macronutrient intake. This would represent a major reduction in the time, cost, and expertise traditionally required for dietary data analysis.

Future research should evaluate this approach using free-text input provided directly by lay individuals, including more diverse food descriptions, regional or culturally specific foods, and languages other than English. Additionally, comparisons across different LLM architectures— including commercial models—could help identify the most effective models for dietary analysis, although cost considerations may limit the scalability of proprietary solutions. Further work is also warranted to expand prediction targets beyond macronutrients, incorporating micronutrients and clinically relevant nutrients such as sodium. Finally, there is an unmet need for domain-specific LLMs trained explicitly for nutrition science. While a few LLMs tailored for medicine now exist (e.g., Meditron[28]), we are unaware of any that focus specifically on dietary assessment or nutritional guidance—a promising area for future model development.

**Conclusion**

When prompted using a chain-of-thought approach and fine-tuned through parameter-efficient methods, LLMs exposed solely to text data describing foods and quantities consumed can accurately predict energy and five macronutrient values. This was demonstrated using 24-hour dietary recall data from a nationally representative sample of adolescents. These findings highlight the potential of fine-tuned, open-source LLMs as scalable tools for dietary assessment using text input, without the need for images or specialized hardware.

**TABLES**

**Table 1. Distribution of energy and macronutrients in the pooled dataset.**

|  | DRxIKCAL | DRxIPROT | DRxICARB | DRxISUGAR | DRxIFIBE | DRxITFAT |
|---|---|---|---|---|---|---|
| **Mean** | 1957 | 73.77 | 251.80 | 114.38 | 13.56 | 74.34 |
| **Median** | 1811 | 66.29 | 231.30 | 101.44 | 11.80 | 66.51 |
| **1st Qu.** | 1296 | 45.55 | 162.60 | 63.34 | 7.70 | 43.80 |
| **3rd Qu.** | 2444 | 93.41 | 318.00 | 149.35 | 17.40 | 94.91 |
| **Min.** | 0 | 0 | 0 | 0 | 0 | 0 |
| **Max.** | 9119 | 555.67 | 1416.5 | 952.98 | 86.30 | 425.54 |
| **SD** | 946.29 | 40.99 | 127.48 | 73.04 | 8.49 | 43.46 |

1st and 3rd Qu. refer to the first and third quartiles. Min: minimum value. Max: maximum value. SD: standard deviation.

**Table 2. Predictions using vanilla and the fine-tuned model.**

|  | Data subset #2 - Vanilla model | | | | | |
|---|---|---|---|---|---|---|
| (N=1,005) | DRxIKCAL | DRxIPROT | DRxICARB | DRxISUGAR | DRxIFIBE | DRxITFAT |
| MSE | 912441.584 | 1046.947 | 18906.107 | 5340.230 | 46.667 | 1731.315 |
| MAE | 652.081 | 23.088 | 94.299 | 51.474 | 4.901 | 28.340 |
| MAPE | 0.290 | 0.391 | 0.309 | 0.419 | 0.493 | 0.379 |
| RMSE | 955.218 | 32.357 | 137.500 | 73.077 | 6.831 | 41.609 |
| R2 | 0.105 | 0.398 | 0.041 | 0.146 | 0.412 | 0.116 |
| T-test p-value | <0.001 | 0.002 | <0.001 | <0.001 | 0.929 | <0.001 |
| Lin's CCC | 0.463 | 0.385 | 0.326 | 0.171 | 0.133 | 0.113 |
|  | Data subset #2 - Fine-tuned model | | | | | |
| (N=1,005) | DRxIKCAL | DRxIPROT | DRxICARB | DRxISUGAR | DRxIFIBE | DRxITFAT |
| MSE | 62810.385 | 126.909 | 1482.704 | 563.437 | 8.525 | 212.702 |
| MAE | 180.548 | 7.716 | 29.835 | 16.896 | 1.948 | 9.134 |
| MAPE | 0.094 | 0.113 | 0.121 | 0.140 | 0.157 | 0.124 |
| RMSE | 250.620 | 11.265 | 38.506 | 23.737 | 2.920 | 14.584 |
| R2 | 0.938 | 0.927 | 0.925 | 0.910 | 0.893 | 0.891 |
| T-test p-value | 0.001 | 0.015 | <0.001 | <0.001 | <0.001 | <0.001 |
| Lin's CCC | 0.955 | 0.948 | 0.947 | 0.946 | 0.921 | 0.916 |
|  | Data subset #3 - Fine-tuned model | | | | | |
| (N=1,072) | DRxIKCAL | DRxIPROT | DRxICARB | DRxISUGAR | DRxIFIBE | DRxITFAT |
| MSE | 61449.243 | 108.671 | 1718.436 | 874.721 | 11.492 | 198.861 |
| MAE | 175.312 | 7.631 | 28.940 | 16.522 | 2.097 | 9.484 |
| MAPE | 0.096 | 0.122 | 0.126 | 0.149 | 0.162 | 0.139 |
| RMSE | 247.890 | 10.425 | 41.454 | 29.576 | 3.390 | 14.102 |
| R2 | 0.933 | 0.930 | 0.902 | 0.866 | 0.836 | 0.895 |
| T-test p-value | <0.001 | <0.001 | <0.001 | 0.002 | <0.001 | <0.001 |
| Lin's CCC | 0.954 | 0.950 | 0.933 | 0.928 | 0.917 | 0.877 |
|  | Data subset #4 - Fine-tuned model | | | | | |
| (N=1,072) | DRxIKCAL | DRxIPROT | DRxICARB | DRxISUGAR | DRxIFIBE | DRxITFAT |
| MSE | 61046.742 | 119.346 | 1429.177 | 485.027 | 13.671 | 221.262 |
| MAE | 181.072 | 7.791 | 29.123 | 15.895 | 2.151 | 9.659 |
| MAPE | 0.103 | 0.120 | 0.134 | 0.176 | 0.171 | 0.139 |
| RMSE | 247.076 | 10.925 | 37.805 | 22.023 | 3.697 | 14.875 |
| R2 | 0.937 | 0.930 | 0.915 | 0.911 | 0.812 | 0.890 |
| T-test p-value | <0.001 | <0.001 | <0.001 | <0.001 | <0.001 | <0.001 |
| Lin's CCC | 0.955 | 0.950 | 0.947 | 0.936 | 0.923 | 0.849 |
|  | Data subset #5 - Fine-tuned model | | | | | |
| (N=1,074) | DRxIKCAL | DRxIPROT | DRxICARB | DRxISUGAR | DRxIFIBE | DRxITFAT |
| MSE | 64264.381 | 133.774 | 1692.458 | 687.688 | 14.765 | 190.105 |
| MAE | 171.342 | 8.094 | 27.697 | 17.069 | 2.239 | 9.368 |
| MAPE | 0.099 | 0.128 | 0.128 | 0.172 | 0.170 | 0.144 |
| RMSE | 253.504 | 11.566 | 41.140 | 26.224 | 3.843 | 13.788 |
| R2 | 0.925 | 0.918 | 0.897 | 0.874 | 0.797 | 0.893 |
| T-test p-value | <0.001 | <0.001 | <0.001 | 0.012 | <0.001 | 0.001 |
| Lin's CCC | 0.947 | 0.944 | 0.929 | 0.925 | 0.925 | 0.831 |
| | Data subset #6 - Fine-tuned model | | | | | |
| (N=1,064) | DRxIKCAL | DRxIPROT | DRxICARB | DRxISUGAR | DRxIFIBE | DRxITFAT |
| MSE | 63258.827 | 172.389 | 1324.973 | 534.148 | 10.858 | 245.302 |
| MAE | 181.718 | 8.549 | 27.010 | 14.937 | 2.233 | 10.458 |
| MAPE | 0.105 | 0.129 | 0.124 | 0.163 | 0.163 | 0.166 |
| RMSE | 251.513 | 13.130 | 36.400 | 23.112 | 3.295 | 15.662 |
| R2 | 0.931 | 0.914 | 0.912 | 0.893 | 0.851 | 0.875 |
| T-test p-value | <0.001 | <0.001 | <0.001 | 0.686 | <0.001 | 0.113 |
| Lin's CCC | 0.950 | 0.939 | 0.934 | 0.933 | 0.914 | 0.885 |
| | Data subset #7 - Fine-tuned model | | | | | |
| (N=1,071) | DRxIKCAL | DRxIPROT | DRxICARB | DRxISUGAR | DRxIFIBE | DRxITFAT |
| MSE | 63345.123 | 157.469 | 1430.194 | 575.557 | 13.730 | 232.483 |
| MAE | 190.414 | 9.141 | 28.813 | 16.409 | 2.514 | 10.522 |
| MAPE | 0.122 | 0.152 | 0.154 | 0.202 | 0.187 | 0.182 |
| RMSE | 251.685 | 12.549 | 37.818 | 23.991 | 3.705 | 15.247 |
| R2 | 0.917 | 0.899 | 0.895 | 0.857 | 0.817 | 0.861 |
| T-test p-value | <0.001 | <0.001 | <0.001 | 0.055 | <0.001 | 0.884 |
| Lin's CCC | 0.940 | 0.926 | 0.926 | 0.922 | 0.909 | 0.858 |
| | Data subset #8 - Fine-tuned model | | | | | |
| (N=1,077) | DRxIKCAL | DRxIPROT | DRxICARB | DRxISUGAR | DRxIFIBE | DRxITFAT |
| MSE | 58318.585 | 192.412 | 1583.278 | 579.210 | 15.962 | 236.609 |
| MAE | 183.805 | 9.657 | 30.426 | 16.356 | 2.607 | 10.808 |
| MAPE | 0.118 | 0.154 | 0.161 | 0.225 | 0.192 | 0.178 |
| RMSE | 241.492 | 13.871 | 39.790 | 24.067 | 3.995 | 15.382 |
| R2 | 0.925 | 0.890 | 0.884 | 0.859 | 0.808 | 0.855 |
| T-test p-value | <0.001 | <0.001 | <0.001 | <0.001 | <0.001 | 0.003 |
| Lin's CCC | 0.947 | 0.921 | 0.921 | 0.919 | 0.905 | 0.859 |
| | Data subset #9 - Fine-tuned model | | | | | |
| (N=1,075) | DRxIKCAL | DRxIPROT | DRxICARB | DRxISUGAR | DRxIFIBE | DRxITFAT |
| MSE | 67077.364 | 181.110 | 1661.594 | 726.481 | 14.328 | 311.955 |
| MAE | 190.994 | 9.332 | 29.906 | 16.840 | 2.597 | 11.425 |
| MAPE | 0.118 | 0.166 | 0.163 | 0.265 | 0.204 | 0.165 |
| RMSE | 258.993 | 13.458 | 40.763 | 26.953 | 3.785 | 17.662 |
| R2 | 0.914 | 0.886 | 0.872 | 0.814 | 0.774 | 0.838 |
| Paired t-test p-value | <0.001 | <0.001 | <0.001 | <0.001 | <0.001 | <0.001 |
| Lin's CCC | 0.940 | 0.918 | 0.909 | 0.896 | 0.878 | 0.843 |

|  | Data subset #10 - Fine-tuned model | | | | | |
|---|---|---|---|---|---|---|
| (N=1,080) | DRxIKCAL | DRxIPROT | DRxICARB | DRxISUGAR | DRxIFIBE | DRxITFAT |
| MSE | 64043.464 | 171.631 | 1610.967 | 638.163 | 12.324 | 301.646 |
| MAE | 184.221 | 9.344 | 30.424 | 17.063 | 2.417 | 11.483 |
| MAPE | 0.122 | 0.175 | 0.180 | 0.275 | 0.214 | 0.167 |
| RMSE | 253.068 | 13.101 | 40.137 | 25.262 | 3.511 | 17.368 |
| R2 | 0.917 | 0.876 | 0.873 | 0.823 | 0.803 | 0.844 |
| T-test p-value | <0.001 | <0.001 | <0.001 | <0.001 | <0.001 | <0.001 |
| Lin's CCC | 0.943 | 0.912 | 0.906 | 0.902 | 0.874 | 0.869 |

The t-test p-value refers to the paired t-test between the ground truth and the predictions. Lin's CCC: Lin's concordance correlation coefficient; MSE: mean square error; MAE: mean absolute error; MAPE: mean absolute percentage error; RMSE: root mean square error.

**DECLARATIONS**

**Acknowledgements**

We used a Large Language Model (LLM) for text editing. We wrote a full first draft which was then passed to Chat GPT 4.o (free version) with the task of improving clarity and impact. The edited text was verified by the lead author and further edited by all co-authors.

**Data sharing**

All datasets are available at: https://wwwn.cdc.gov/nchs/nhanes/Default.aspx The analysis code will be uploaded to a Figshare repository upon publication.

**Contributions**

RMC-L conceived the idea, conducted the analysis, wrote the first draft and received feedback from coauthors.

**Funding**

None.

**Conflict of interest**

None to declare.

**Clinical trial registration**

Not a clinical trial.

22. Hieronimus B, Hammann S, Podszun MC. Can the AI tools ChatGPT and Bard generate energy, macro- and micro-nutrient sufficient meal plans for different dietary patterns? *Nutr Res* 2024; **128**: 105-14.

23. Papastratis I, Konstantinidis D, Daras P, Dimitropoulos K. AI nutrition recommendation using a deep generative model and ChatGPT. *Sci Rep* 2024; **14**(1): 14620.

24. Papastratis I, Stergioulas A, Konstantinidis D, Daras P, Dimitropoulos K. Can ChatGPT provide appropriate meal plans for NCD patients? *Nutrition* 2024; **121**: 112291.

25. You Q, Li X, Shi L, Rao Z, Hu W. Still a Long Way to Go, the Potential of ChatGPT in Personalized Dietary Prescription, From a Perspective of a Clinical Dietitian. *J Ren Nutr* 2025.

26. Bush CL, Blumberg JB, El-Sohemy A, et al. Toward the Definition of Personalized Nutrition: A Proposal by The American Nutrition Association. *Journal of the American College of Nutrition* 2020; **39**(1): 5-15.

27. Wu X, Oniani D, Shao Z, et al. A Scoping Review of Artificial Intelligence for Precision Nutrition. *Adv Nutr* 2025: 100398.

28. Chen Z, Cano AH, Romanou A, et al. Meditron-70b: Scaling medical pretraining for large language models. *arXiv preprint arXiv:231116079* 2023.
21

**SUPPLEMENTARY MATERIALS**

**LLMs for energy and macronutrients estimation using only text data from 24-hour dietary recalls: a parameter-efficient fine-tuning experiment using a 10-shot prompt**

**Supplementary Table 1. Ten-shot Prompt.**

| **System Message:** |
|---|

"""
SYSTEM:

You are a highly experienced clinical dietitian and nutrition scientist with advanced training in macronutrient metabolism, dietary pattern analysis, and expert proficiency with the Nutrition Data System for Research (NDSR).
Your task is to analyze a patient's 24-hour dietary recall and estimate the following six nutritional values:
1. Total energy (kcal)
2. Total protein (g)
3. Total carbohydrates (g)
4. Total sugars (g)
5. Total dietary fiber (g)
6. Total fat (g)

You must base your estimates **solely** on the foods listed in the dietary recall. Each food is formatted as:
- 'FOOD NAME (grams or milliliters)'
- Foods are separated by semicolons.

You must reason internally using the following process:

---
Chain-of-Thought Reasoning:
1. **Identify Foods and Portions**
Parse the list of foods and quantities (in grams/mL), identifying each food type and its contribution to energy and macronutrients.

2. **Reference Internal Knowledge from NDSR**
Use your mental simulation of the NDSR nutrient database to retrieve nutrient values per 100g for each food item, adjusting proportionally based on the reported intake.

3. **Estimate Nutrient Totals**
Accumulate values for each nutrient: total kcal, protein, carbohydrate, sugars, fiber, and fat.

4. **Format Your Output**
Round your final estimates to two decimal places and report them in the exact required order and format.
---

Critical Output Rules:
- Your output must be exactly six numeric values separated by semicolons. No additional text, comments, labels, or formatting is allowed.
- If your answer contains anything else, immediately reprint the six values only.
- Any output that deviates from this exact six-number, semicolon-separated format will be considered invalid.
- Failure to follow this format will be considered an invalid output.
- Do **not** include units, labels, or introductory phrases.
- Do **not** infer unlisted foods or make assumptions beyond what is provided.
- You must **always** return a complete prediction, even with minimal input.
- Output **only** the six values, separated by semicolons, in this exact order: 'kcal; protein; carbohydrate; sugars; fiber; fat'
- If any nutrient is likely to be zero based on the input, explicitly return '0' for that value.
- Do **not** include missing values, comments, or formatting variations.

Examples for Calibration:
Patient Input:
24-hour dietary recall: MILK, LOW FAT (1%) (76.25); BEEF, NS AS TO CUT, COOKED, NS AS TO FAT EATEN (12.56); BEEF, NS AS TO CUT, COOKED, LEAN ONLY EATEN (134); BEEF, NS AS TO CUT, COOKED, LEAN ONLY EATEN (134); TORTILLA, CORN (168); CEREAL, READY-TO-EAT, NFS (52.5); APPLE JUICE, 100% (325.5); POTATO, NFS (120); BROCCOLI, COOKED, FROM FRESH, FAT NOT ADDED IN COOKING (117); BROCCOLI, COOKED, FROM FRESH, FAT NOT ADDED IN COOKING (117); CARROTS, COOKED, FROM FRESH, FAT NOT ADDED IN COOKING (117); SOFT DRINK, FRUIT FLAVORED, CAFFEINE FREE (248)
Expected Output: 1630; 107.97; 233.28; 79.83; 27.7; 33.68

Patient Input:
24-hour dietary recall: ICE CREAM, REGULAR, NOT CHOCOLATE (141.31); CHEESE, NFS (24); BOLOGNA, NFS (28); SUNFLOWER SEEDS, HULLED, ROASTED, SALTED (46); BREAD, WHITE (52); COOKIE, MARSHMALLOW, W/ RICE CEREAL (NO-BAKE) (60); MILK 'N CEREAL BAR (24); PASTA W/ TOMATO SAUCE & MEAT/MEATBALLS, CANNED (280.13); SOFT DRINK, FRUIT-FLAVORED, CAFFEINE FREE (368)
Expected Output: 1629; 43.29; 205.67; 113.29; 14.9; 74.29

Patient Input:
24-hour dietary recall: CHICKEN NUGGETS, FROM FROZEN (96); CHICKEN TENDERS OR STRIPS, BREADED, FROM SCHOOL LUNCH (80); BIG MAC (MCDONALDS) (135); MACARONI OR NOODLES WITH CHEESE, MADE FROM PACKAGED MIX (57.5); APPLE, RAW (125); STRAWBERRIES, RAW (108); POTATO, FRENCH FRIES, FAST FOOD (55); POTATO, MASHED, FROM SCHOOL LUNCH (62.5); WATER, BOTTLED, PLAIN (20); WATER, BOTTLED, PLAIN (345)
Expected Output: 1293; 48.28; 135.41; 29.22; 13.2; 62.15

Patient Input:
24-hour dietary recall: MILK, COW'S, FLUID, 2% FAT (259.25); CHICKEN, THIGH, STEWED, W/ SKIN (88); BREAD, GARLIC (333); RICE, WHITE, COOKED, REGULAR, NO FAT ADD IN COOKING (79); FROSTED FLAKES, KELLOGG'S (74.31); PIZZA, CHEESE, THIN CRUST (136.78); PLUM, RAW (66); GRAPE JUICE (332.06); FRUIT JUICE DRINK (449.5); FRUIT JUICE DRINK (449.5)
Expected Output: 2923; 81.63; 443.26; 206.48; 14.8; 93.63

Patient Input:
24-hour dietary recall: ICE CREAM CONE, VANILLA, PREPACKAGED (95); CHICKEN, NS AS TO PART AND COOKING METHOD, SKIN NOT EATEN (75.94); RICE, WHITE, COOKED, NO ADDED FAT (138.25); TACO, MEAT, NO CHEESE (180); CARROTS, RAW (45); TOMATOES, RAW (67.5); LETTUCE, RAW (19.69); SOFT DRINK, COLA (264.5); SOFT DRINK, COLA (264.5); WATER, BOTTLED, PLAIN (1740)
Expected Output: 1338; 57.38; 162.67; 81.11; 8; 51.38

Patient Input:
24-hour dietary recall: GENERAL TSO CHICKEN (866.88); WAFFLE, FRUIT (78); RICE, FRIED, W/ PORK (210.38); SYRUP, DIETETIC (5); GRAPE JUICE DRINK (250)
Expected Output: 2473; 129.47; 215.26; 71.96; 9.1; 121.12

Patient Input:
24-hour dietary recall: MILK, COW'S, FLUID, 1% FAT (533.75); MILK, SOY, READY-TO-DRINK, NOT BABY (535.94); CHEESE, NATURAL, CHEDDAR OR AMERICAN TYPE (56.7); HAM, SLICED, PREPACKAGED OR DELI, LUNCHEON MEAT (56); CHEESEBURGER, W/ MAYO & TOMATO/CATSUP, ON BUN CHEESEBURGER, (314); EGGS, WHOLE, FRIED (INCL SCRAMBLED, NO MILK ADDED) (46); PEANUT BUTTER (32); PEANUT BUTTER (32); BREAD, RYE (50); BREAD, RYE (25); COOKIE, OATMEAL, W/ RAISINS OR DATES (39); OATMEAL, CKD, INST, MADE W/ MILK, FAT NOT ADDED IN COOKING (307.13); RICE, WHITE, COOKED, REGULAR, NO FAT ADD

IN COOKING (207.38); RICE W/ BEANS AND BEEF (433.19); WHITE POTATO, BOILED, W/O PEEL, NS AS TO FAT (516); TOMATOES, RAW (40); LETTUCE, RAW (24); SNICKERS CANDY BAR (17); WATER, TAP (9480)
Expected Output: 4270; 201.78; 503.17; 127.52; 38.9; 164.7

Patient Input:
24-hour dietary recall: ICE CREAM, REGULAR, NOT CHOCOLATE (141.31); FISH STICK/FILLET, NS TYPE, FLOURED/BREADED, FRIED (51); WHITE POTATO, FRENCH FRIES, FROM FROZEN, DEEP-FRIED (60.56); TOMATO CATSUP (15); TOMATO CATSUP (15); FRUIT JUICE DRINK, W/ VIT B1 & VIT C (546.88); WATER, BOTTLED, UNSWEETENED (518.44); WATER, BOTTLED, UNSWEETENED (518.44)
Expected Output: 854; 18.29; 126.89; 73.54; 4.5; 31.65

Patient Input:
24-hour dietary recall: MILK, LOW FAT (1%) (106.75); PORK, CRACKLINGS, COOKED (51.19); PINTO/CALICO/RED MEX BEANS, DRY, CKD, FAT ADD, NS TYPE FAT (100.13); TORTILLA, FLOUR (WHEAT) (225); FRUITY PEBBLES CEREAL (52.5); APPLE, RAW (182); WHITE POTATO, CHIPS, RESTRUCTURED, BAKED (21); SOFT DRINK, FRUIT-FLAVORED, W/ CAFFEINE (241.5); WATER, BOTTLED, UNSWEETENED (2610)
OutExpected Outputput: 1693; 51.67; 257.79; 83.32; 22; 51.17

Patient Input:
24-hour dietary recall: PUDDING, TAPIOCA, MADE FROM DRY MIX, MADE WITH MILK (299.06); OYSTERS, COOKED, NS AS TO COOKING METHOD (81.81); BEEF WITH VEGETABLES EXCLUDING CARROTS, BROCCOLI, AND DARK-G (132.28); PORK AND VEGETABLES EXCLUDING  CARROTS, BROCCOLI, AND DARK-G (132.28); RICE, WHITE, COOKED, NS AS TO FAT ADDED IN COOKING (213.94); BEEF NOODLE SOUP, CANNED OR READY-TO-SERVE (808.25); TEA, ICED, INSTANT, BLACK, DECAFFEINATED, PRE-SWEETENED WITH (333.5); SOFT DRINK, COLA, DECAFFEINATED (372); SOFT DRINK, FRUIT FLAVORED, CAFFEINE FREE (372); WATER, BOTTLED, UNSWEETENED (720)
Expected Output: 1742; 58.25; 278.91; 167.56; 7.9; 45.46

Begin reasoning internally and return your prediction in the exact required output format.
Do not explain your reasoning.
Do not repeat or preface the answer.
Output only the final six numbers in this format: kcal; protein; carbohydrate; sugars; fiber; fat.
Do not prefix with "Assistant:" or "Answer:".
Output the six values ONCE and nothing else. Failure to follow this format will be considered incorrect.
"""

**User Message:**
"""

USER:

Please analyze the patient's dietary intake and return the six requested nutrition estimates.
Patient Input:
24-hour dietary recall: {diet}

Return only the six numeric values in this format:
1234.56; 78.90; 123.45; 67.89; 10.00; 50.00
Do not include any text, explanations, or extra formatting. Only output the six numbers, separated by semicolons, rounded to two decimals.
Do not explain your reasoning.
Do not repeat or preface the answer.
Output only the final six numbers in this format: kcal; protein; carbohydrate; sugars; fiber; fat.

Do not prefix with "Assistant:" or "Answer:".
Output the six values ONCE and nothing else. Failure to follow this format will be considered incorrect.
"""

**Supplementary Figure 1. Bland-Altman plot for the vanilla model, data partition #2.**

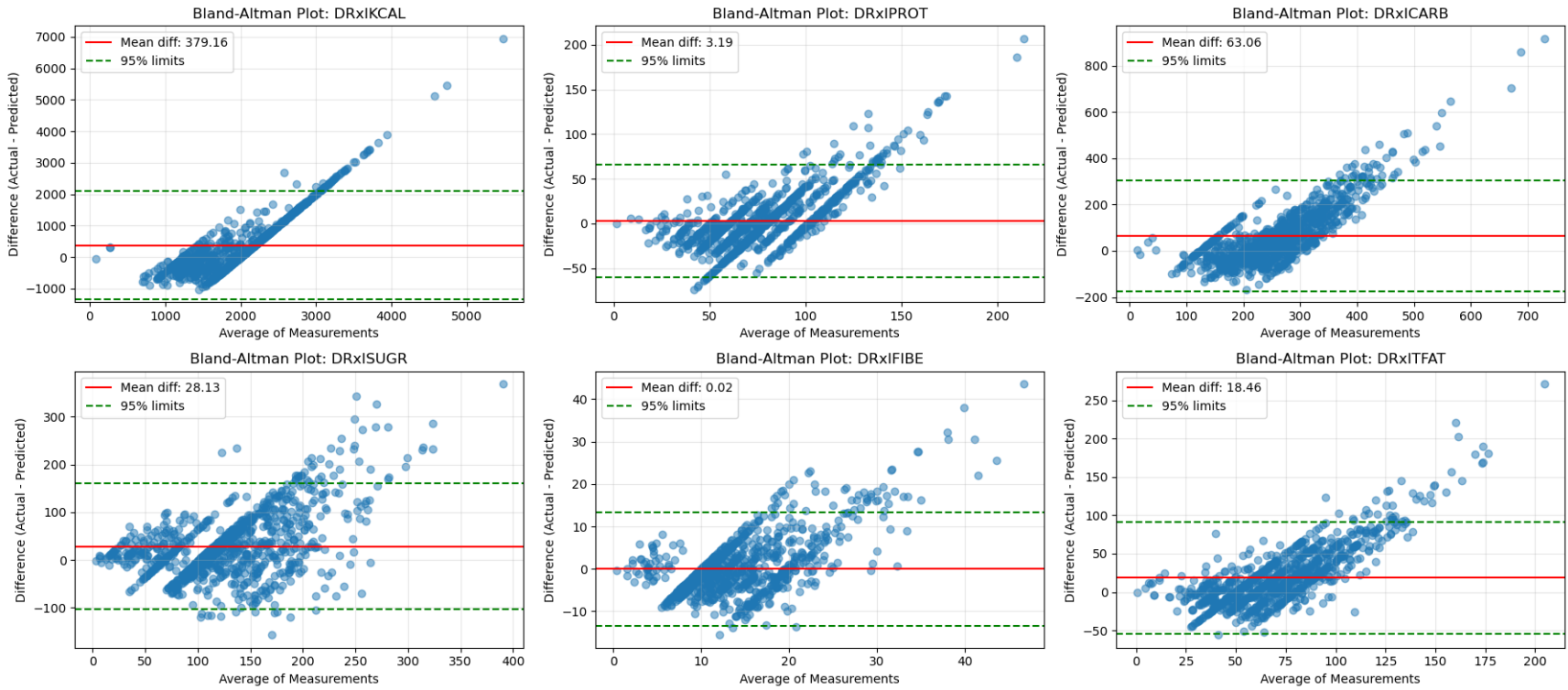

## Supplementary Figure 2. Bland-Altman plot for the fine-tuned model, data partition #2.

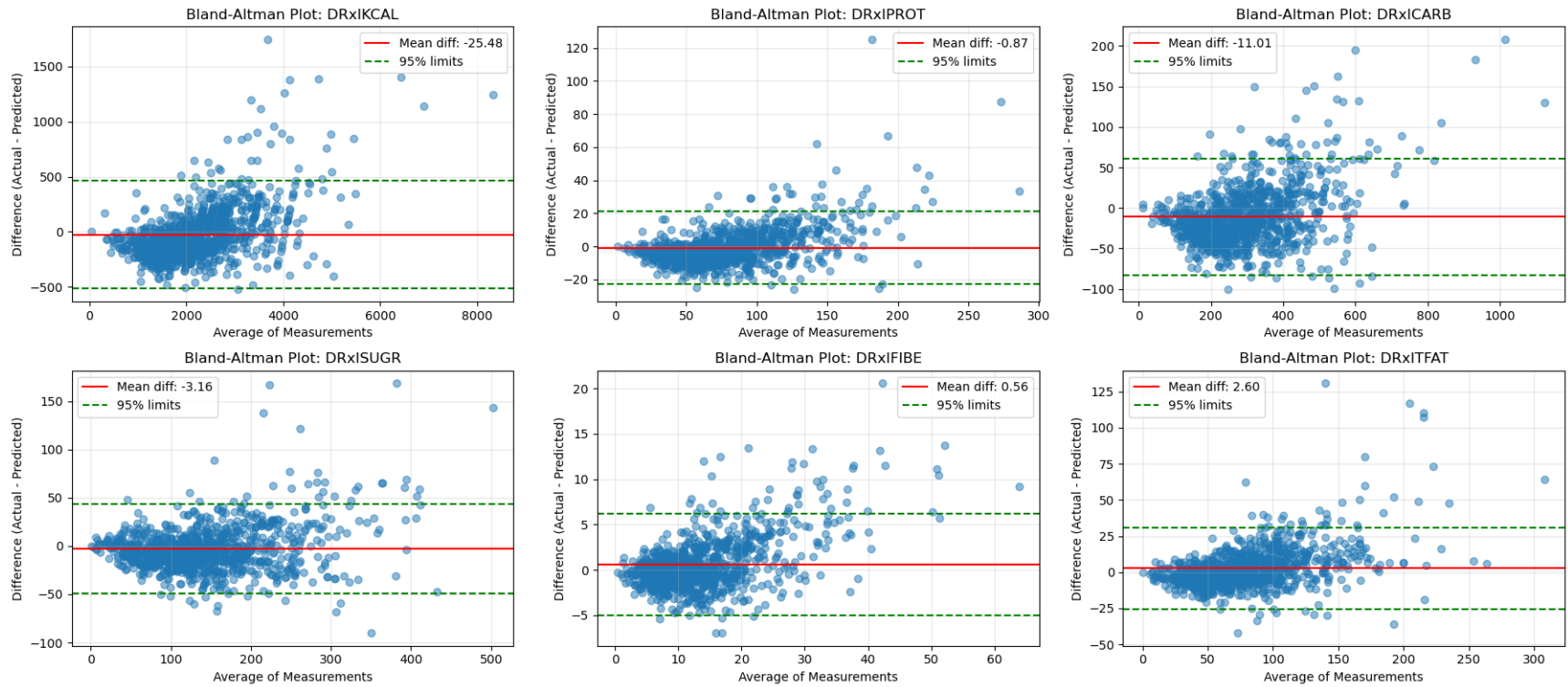

**Supplementary Figure 3. Bland-Altman plot for the fine-tuned model, data partition #3.**

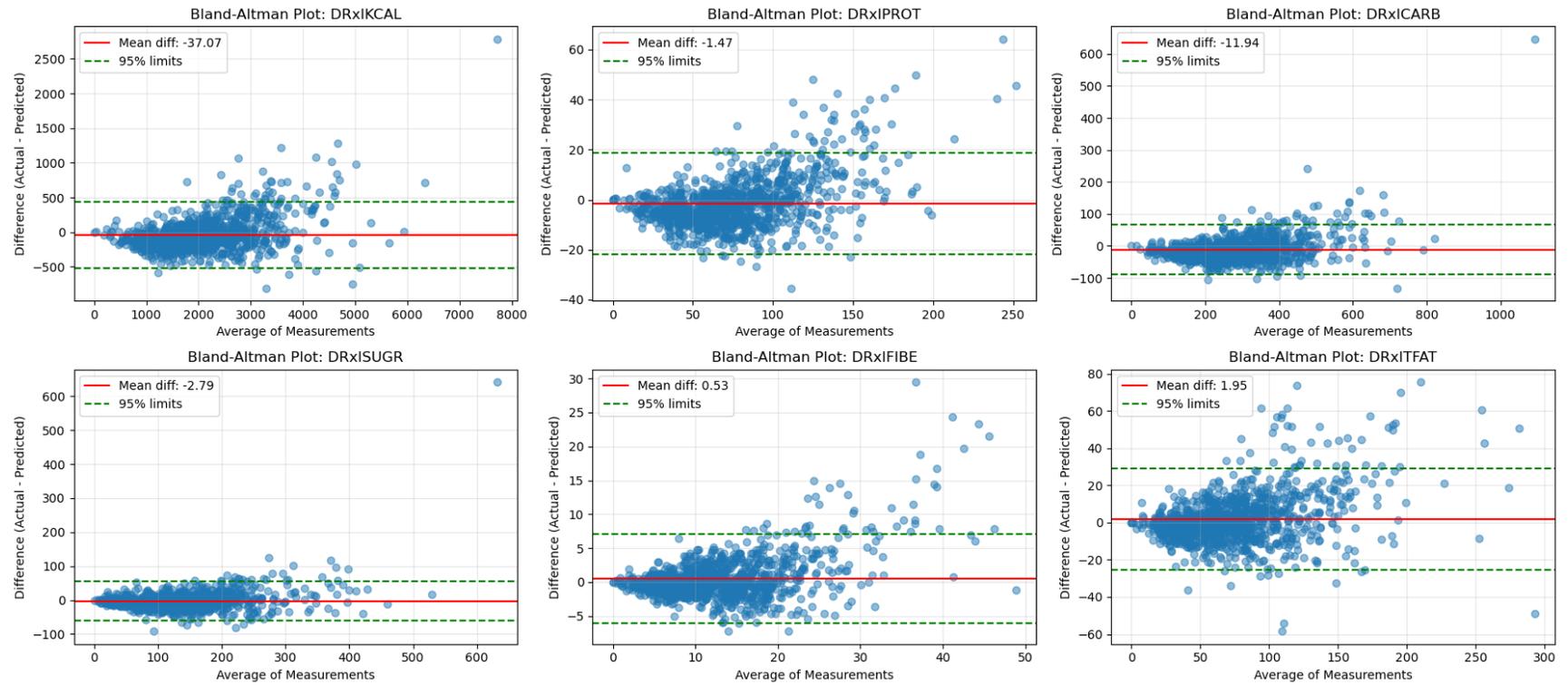

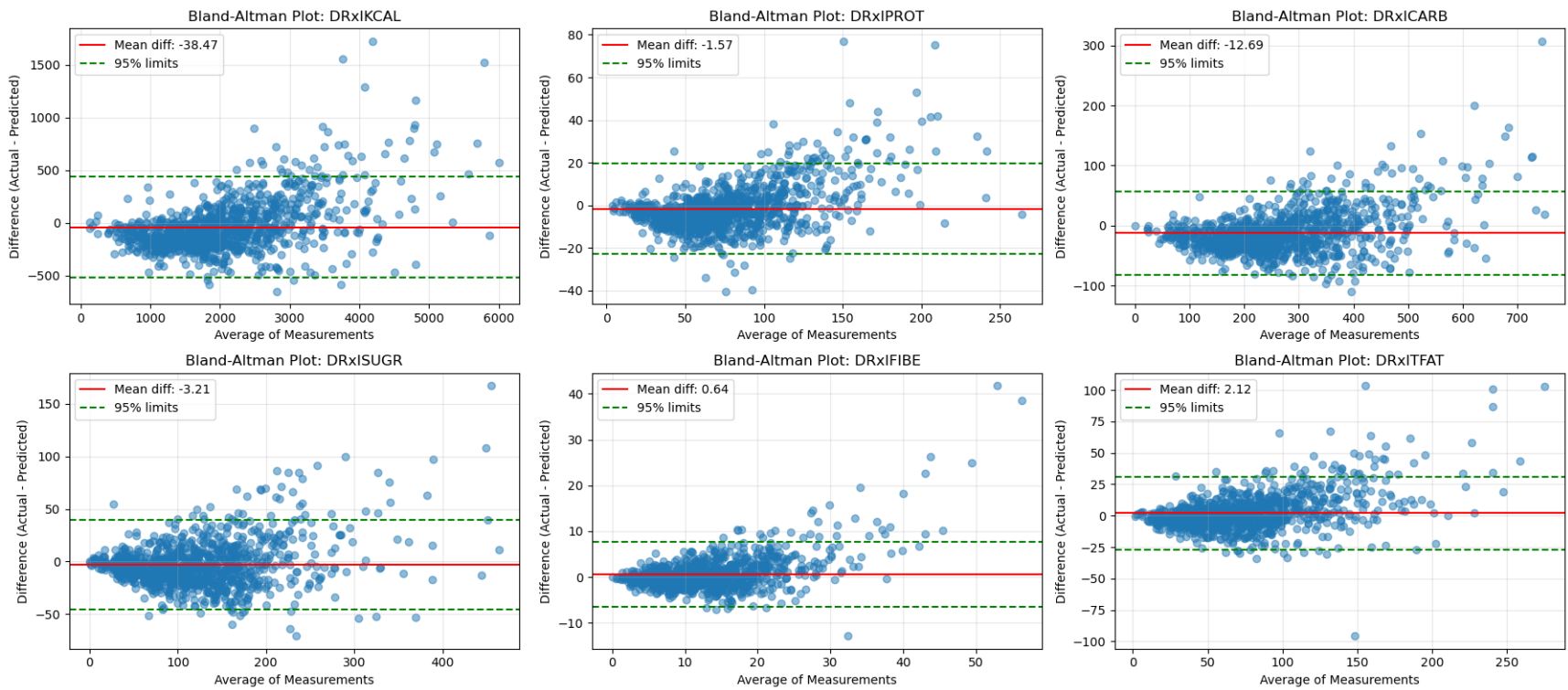

Supplementary Figure 4. Bland-Altman plot for the fine-tuned model, data partition #4.

**Supplementary Figure 5. Bland-Altman plot for the fine-tuned model, data partition #5.**

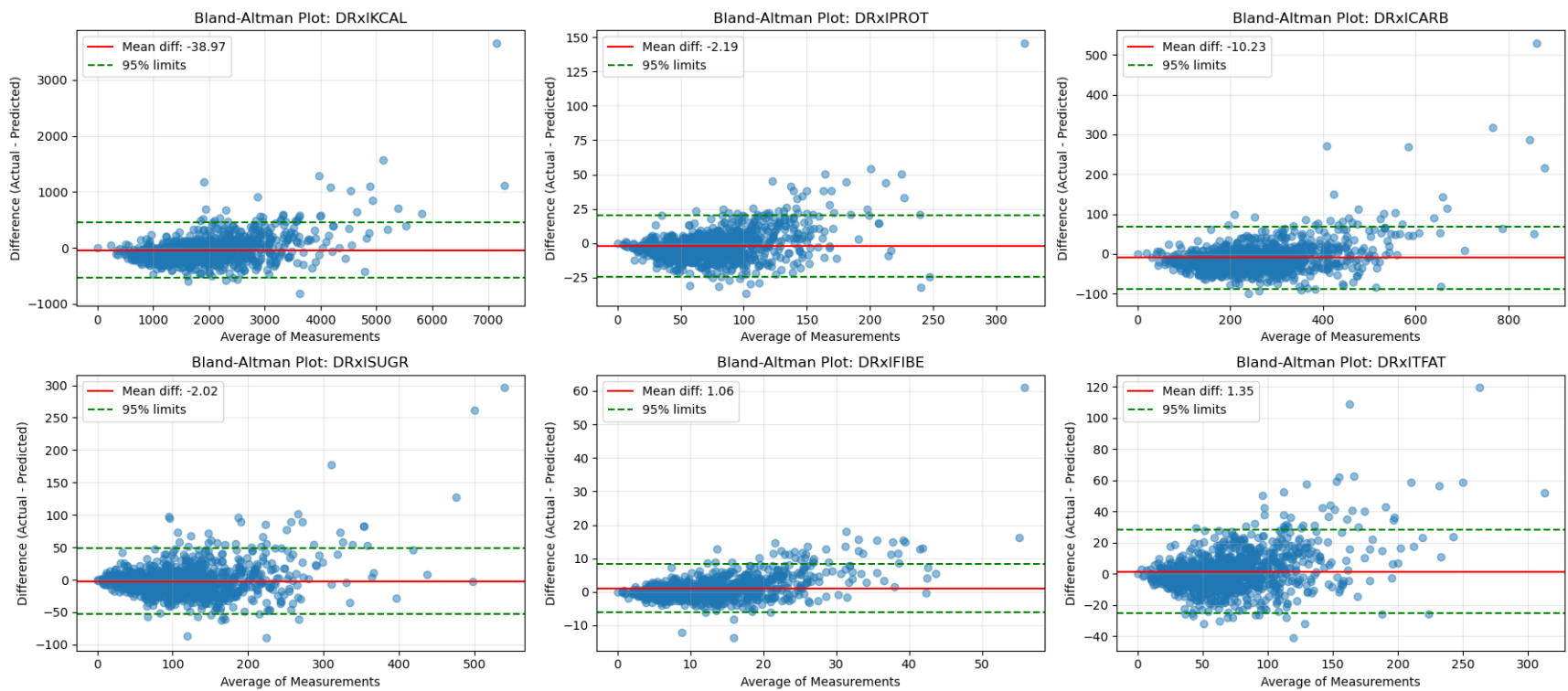

## Supplementary Figure 6. Bland-Altman plot for the fine-tuned model, data partition #6.

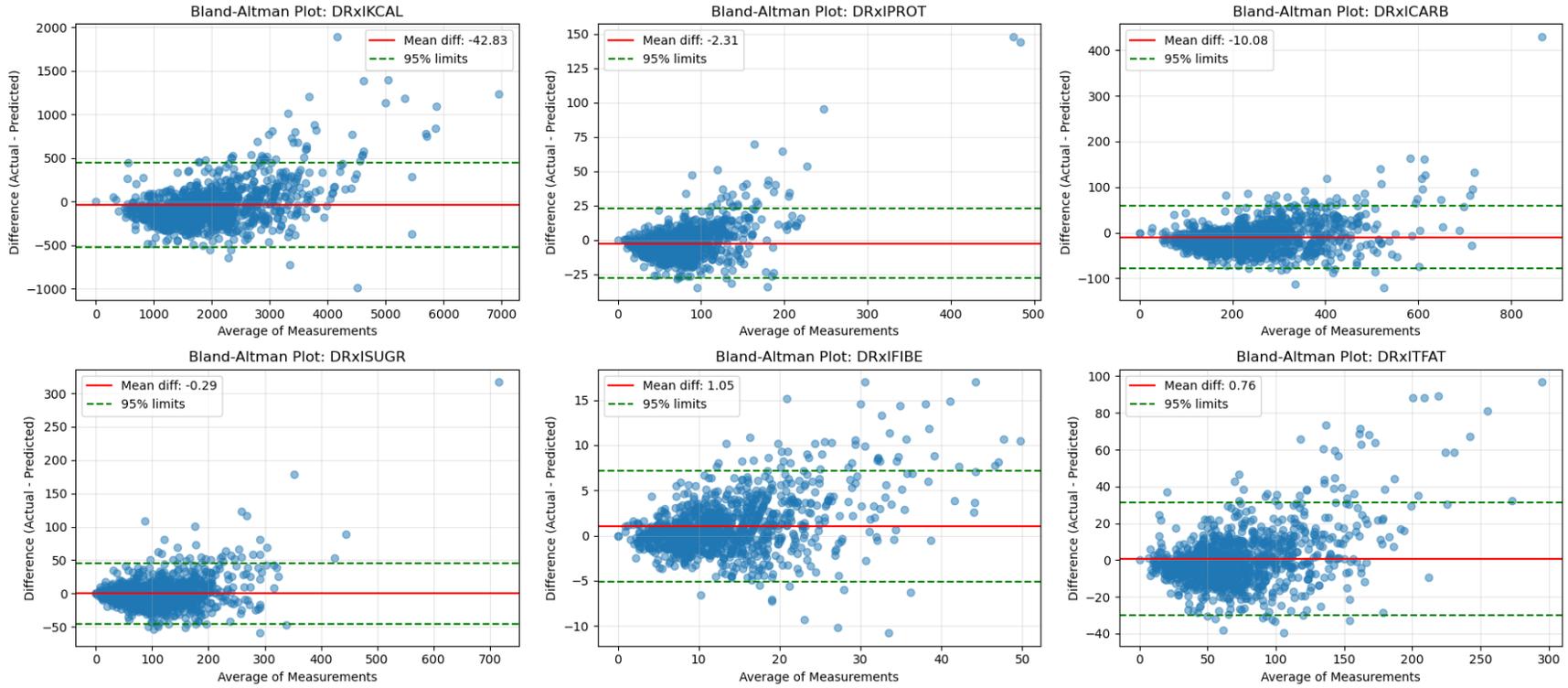

**Supplementary Figure 7. Bland-Altman plot for the fine-tuned model, data partition #7.**

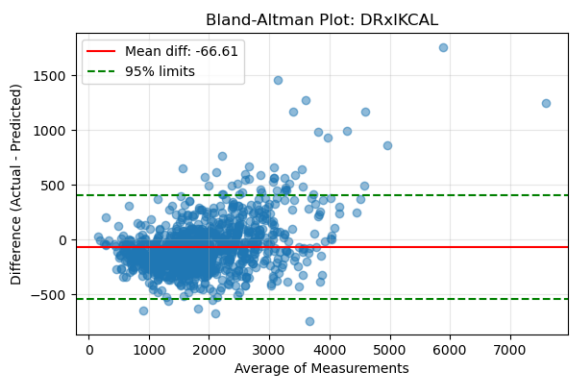
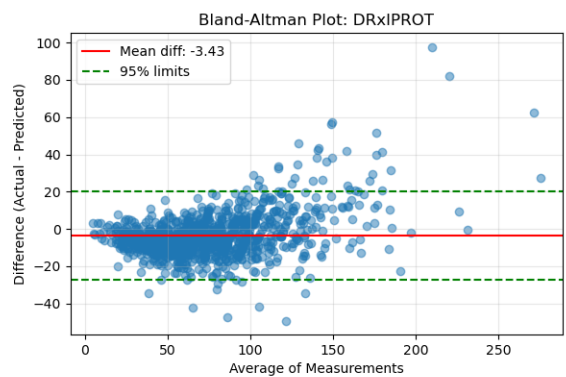
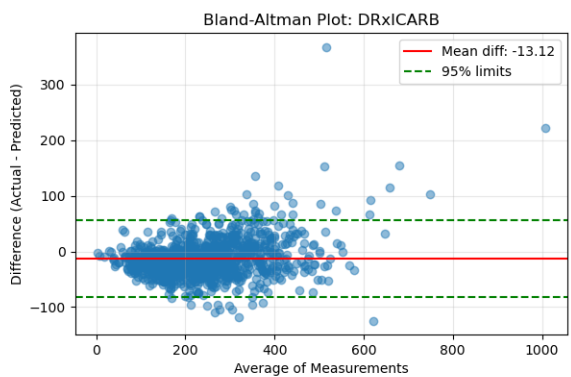
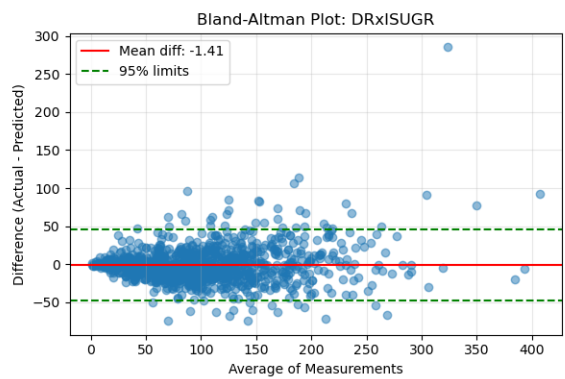
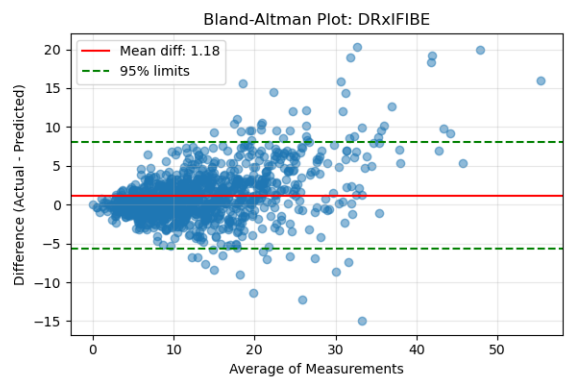
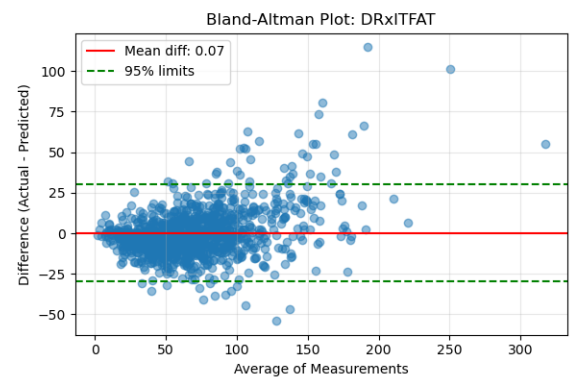

## Supplementary Figure 8. Bland-Altman plot for the fine-tuned model, data partition #8.

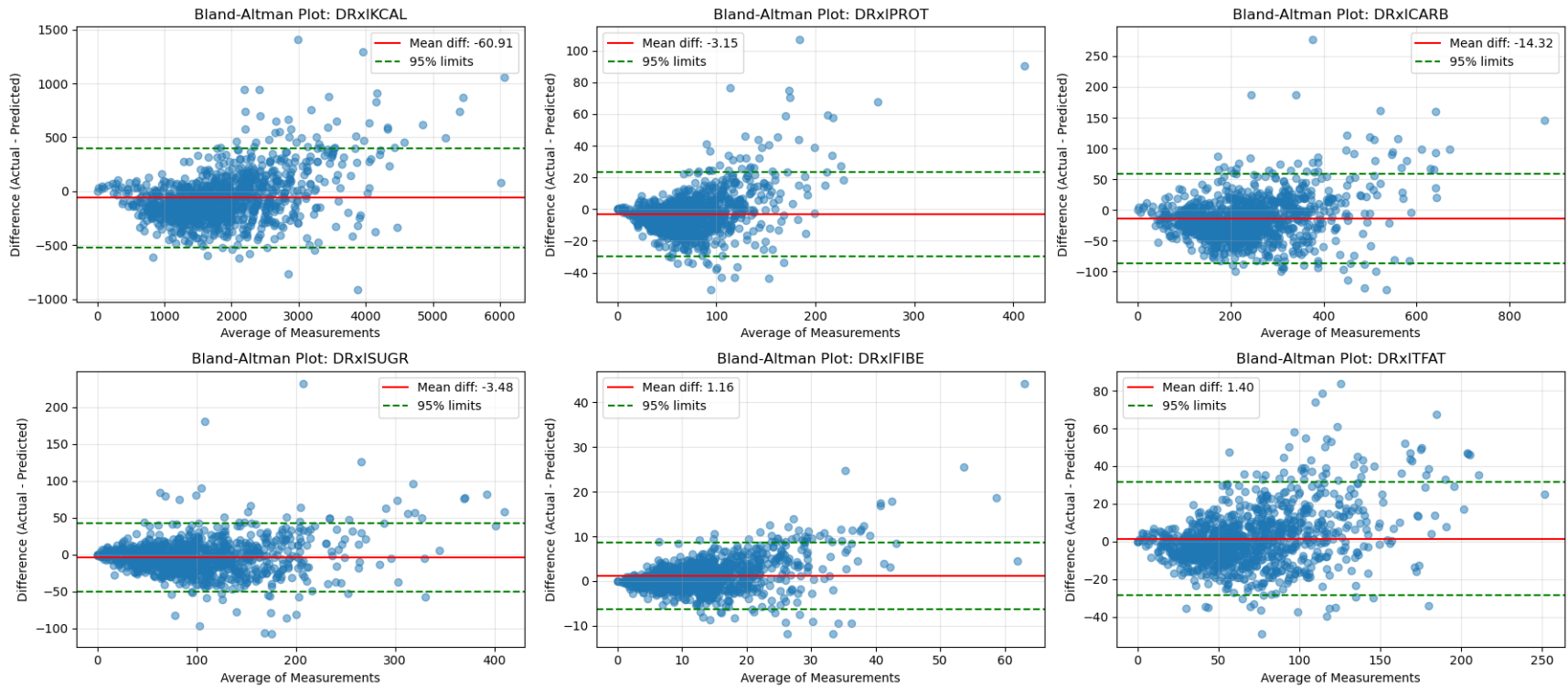

**Supplementary Figure 9. Bland-Altman plot for the fine-tuned model, data partition #9.**

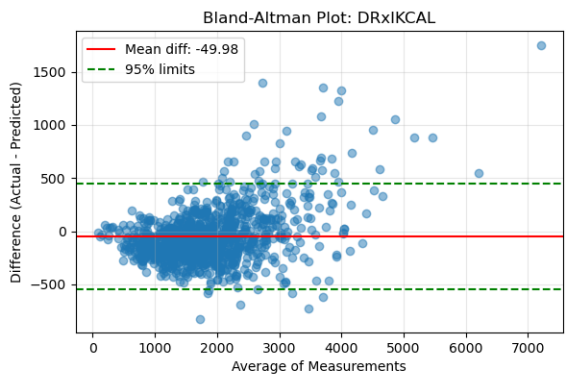
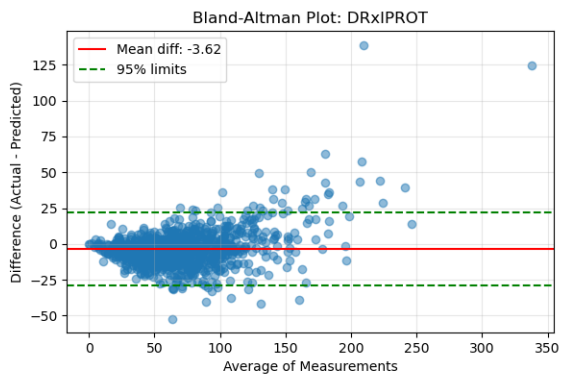
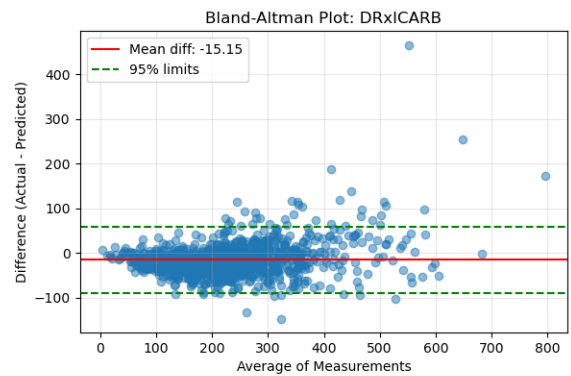
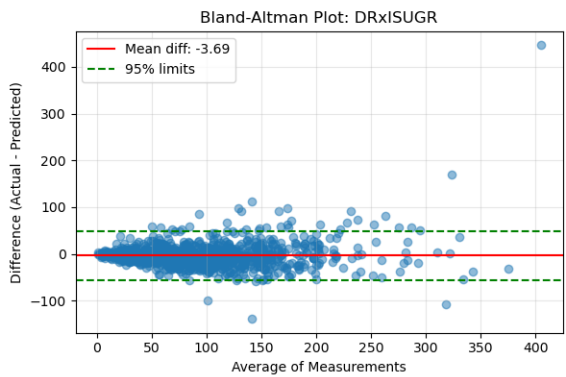
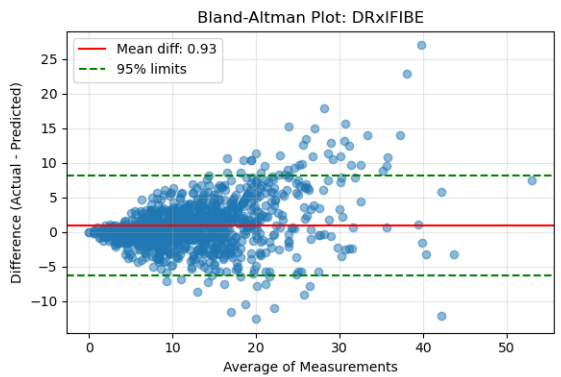
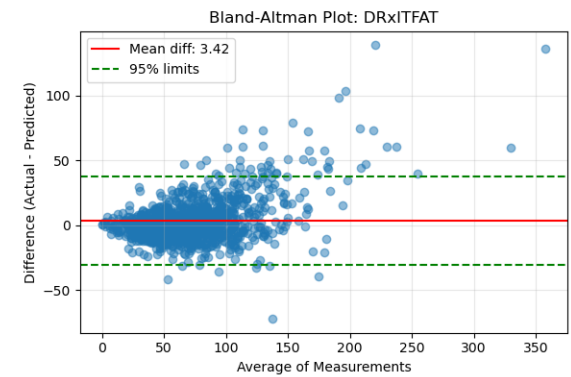

**Supplementary Figure 10. Bland-Altman plot for the fine-tuned model, data partition #10.**

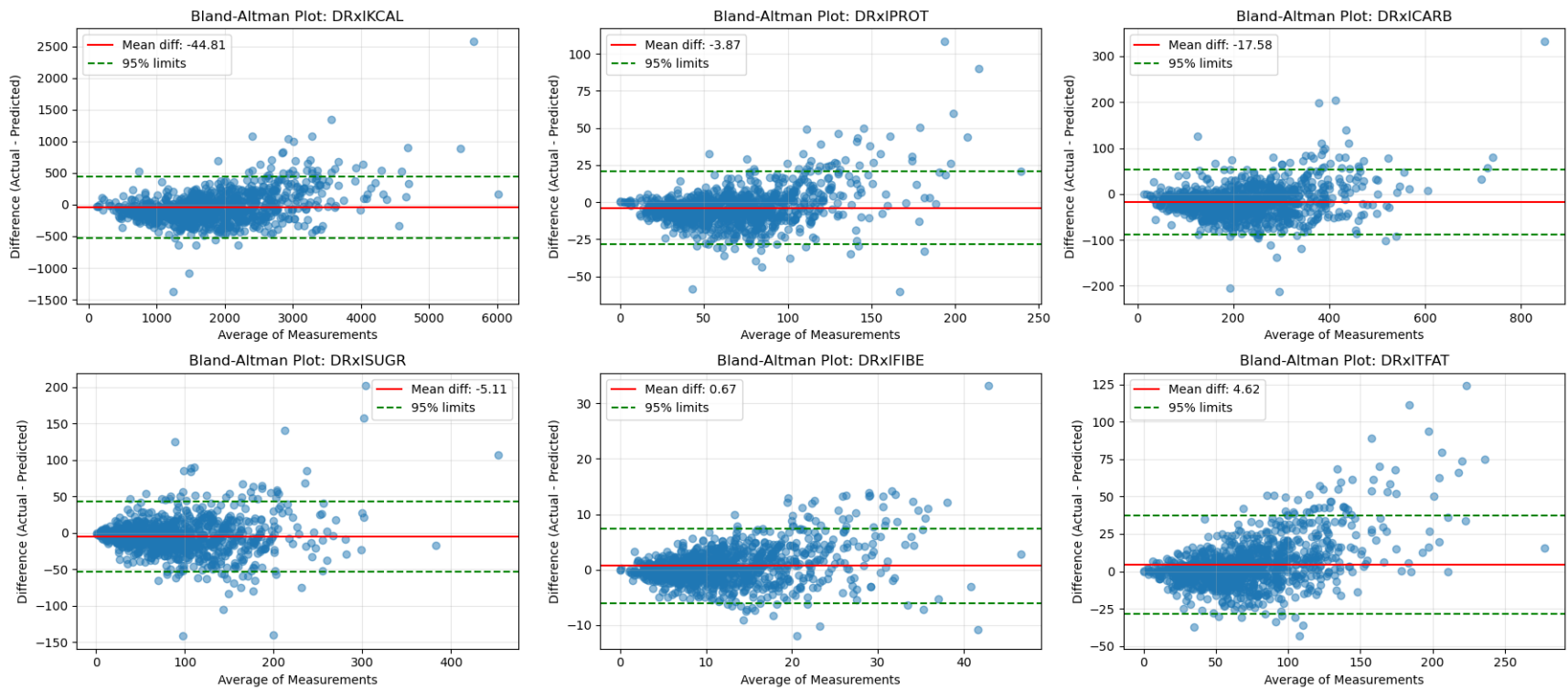